\documentclass[letterpaper, 10 pt, conference]{ieeeconf}  

\IEEEoverridecommandlockouts 
\usepackage{graphicx}
\usepackage{amsmath}
\usepackage{amssymb}
\usepackage{booktabs}
\usepackage{color, colortbl}
\usepackage{booktabs}
\usepackage[percent]{overpic}
\usepackage{xcolor}
\usepackage{verbatim}

\usepackage{multirow}

\usepackage{subcaption}
\usepackage{balance}

\usepackage[pagebackref,breaklinks,colorlinks]{hyperref}

\usepackage[capitalize]{cleveref}
\crefname{section}{Sec.}{Secs.}
\Crefname{section}{Section}{Sections}
\Crefname{table}{Table}{Tables}
\crefname{table}{Tab.}{Tabs.}

\definecolor{somegray}{rgb}{0.5, 0.5, 0.5}
\newcommand{\darkgrayed}[1]{\textcolor{somegray}{#1}}
\makeatletter
\newcommand*\titleheader[1]{\gdef\@titleheader{#1}}
\AtBeginDocument{%
  \let\st@red@title\@title
  \def\@title{%
    \vskip-3em
    \bgroup\normalfont\large\centering\@titleheader\par\egroup
    \vskip1.5em\st@red@title}
}
\makeatother

\titleheader{\darkgrayed{This paper has been accepted for publication at the \\
IEEE/RSJ International Conference on Intelligent Robots and Systems (IROS), Kyoto, 2022.
\copyright IEEE}}

\title{\LARGE \bf
Multi-View Guided Multi-View Stereo
}

\author{Matteo Poggi$^*$\thanks{$^*$joint first authorship}, Andrea Conti$^*$ and Stefano Mattoccia \\
University of Bologna%
}

\begin{document}

\maketitle

\begin{abstract}
This paper introduces a novel deep framework for dense 3D reconstruction from multiple image frames, leveraging a sparse set of depth measurements gathered jointly with image acquisition.
Given a deep multi-view stereo network, our framework uses sparse depth hints to guide the neural network by modulating the plane-sweep cost volume built during the forward step, enabling us to infer constantly much more accurate depth maps.
Moreover, since multiple viewpoints can provide additional depth measurements, we propose a multi-view guidance strategy that increases the density of the sparse points used to guide the network, thus leading to even more accurate results. We evaluate our Multi-View Guided framework within a variety of state-of-the-art deep multi-view stereo networks, demonstrating its effectiveness at improving the results achieved by each of them on BlendedMVG and DTU datasets.
\end{abstract}

\section{Introduction}
\label{sec:intro}

Multi-view stereo (MVS) is a popular technique to obtain dense 3D reconstructions of real-world objects or scenes from a set of multiple, posed images. It represents a first, pivotal step towards a variety of higher-level applications, such as augmented/virtual reality, robotics, cultural heritage and more.
It represents one of the fundamental problems in computer vision and it has been studied for years, at first by developing classical algorithms
\cite{barnes2009patchmatch,campbell2008using,furukawa2009accurate,galliani2015massively,schonberger2016pixelwise}, making use of hand-crafted matching functions to measure consistency among the multiple views. However, many challenges keep MVS an open problem, such as occlusions between the views, lack of texture or non-Lambertian surfaces, to name a few \cite{aanaes2016large,knapitsch2017tanks,schops2017multi}. 

The advent of deep learning in computer vision, in particular with the introduction of Convolutional Neural Networks (CNNs), allowed for rapid progress even in geometric tasks such as MVS, partially overcoming some of the issues mentioned above. Indeed, deep MVS networks \cite{yao2018mvsnet,yao2019recurrent,Wei_2021_ICCV,wang2021patchmatchnet,gu2020cascade,luo2019p} are spreading, thanks to their ever-increasing accuracy on popular benchmarks \cite{jensen2014large,schops2017multi,knapitsch2017tanks}.
Common to most CNNs developed for this purpose is the presence of a 3D cost volume \cite{yao2018mvsnet}, built using plane-sweeping over the source views features and computing their similarity with respect to the reference image features. Such a volume is usually regularized through 3D convolutional layers -- or other, more efficient alternatives, such as 2D Long-Short Term Memory (LSTM) layers \cite{yao2019recurrent} -- before regressing the final depth map. 
However, despite the more robust features representation extracted by 2D CNNs and the strong regularization achieved through 3D convolutions, the high-demanding computational requirements still limit the full deployment of such solutions, often requiring some trade-off between accuracy and complexity. For instance, inferring depth at resolution lower than the one of the input images \cite{yao2018mvsnet} or implementing coarse-to-fine strategies \cite{gu2020cascade,yang2020cost,cheng2020deep}. 
Moreover, several challenges mentioned above, such as dealing with untextured regions, thin objects or occlusions, remain open.

\begin{figure}[t]
    \centering
    \renewcommand{\tabcolsep}{1pt}
    \scalebox{0.9}{
    \begin{tabular}{ccc}
        \textit{RGB (and hints)} & \textit{Estimated depth} & \textit{Point cloud} \\
         \includegraphics[height=0.24\linewidth]{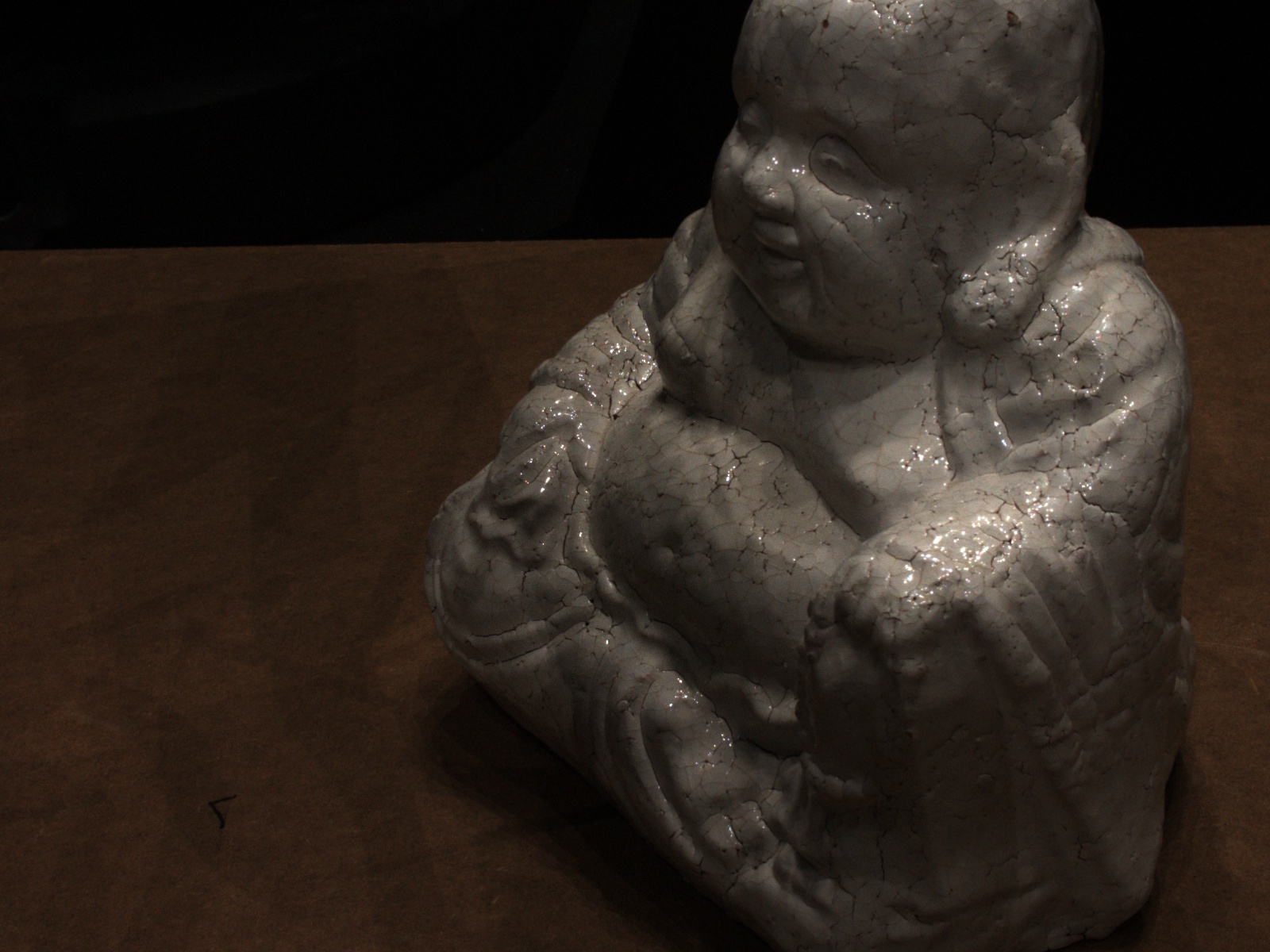} &
         \includegraphics[height=0.24\linewidth]{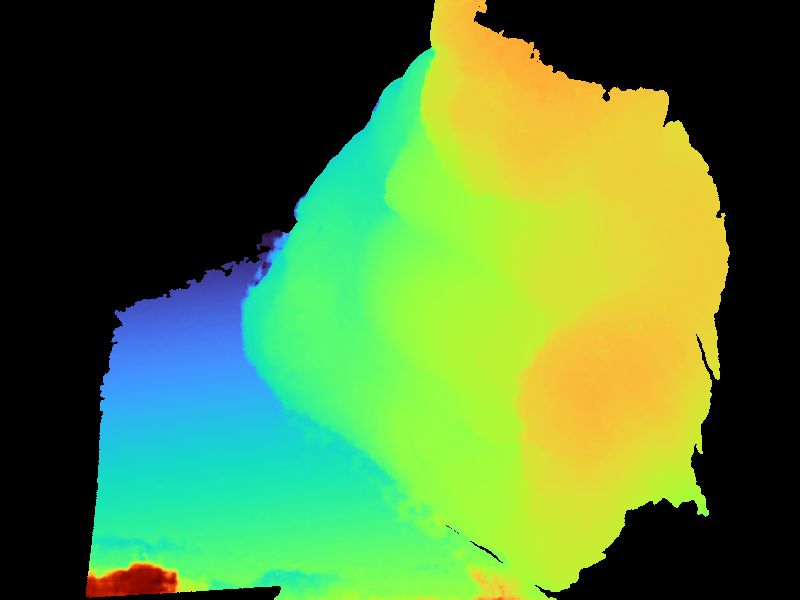} &
         \includegraphics[trim=22cm 12cm 15cm 10cm,clip,height=0.24\linewidth]{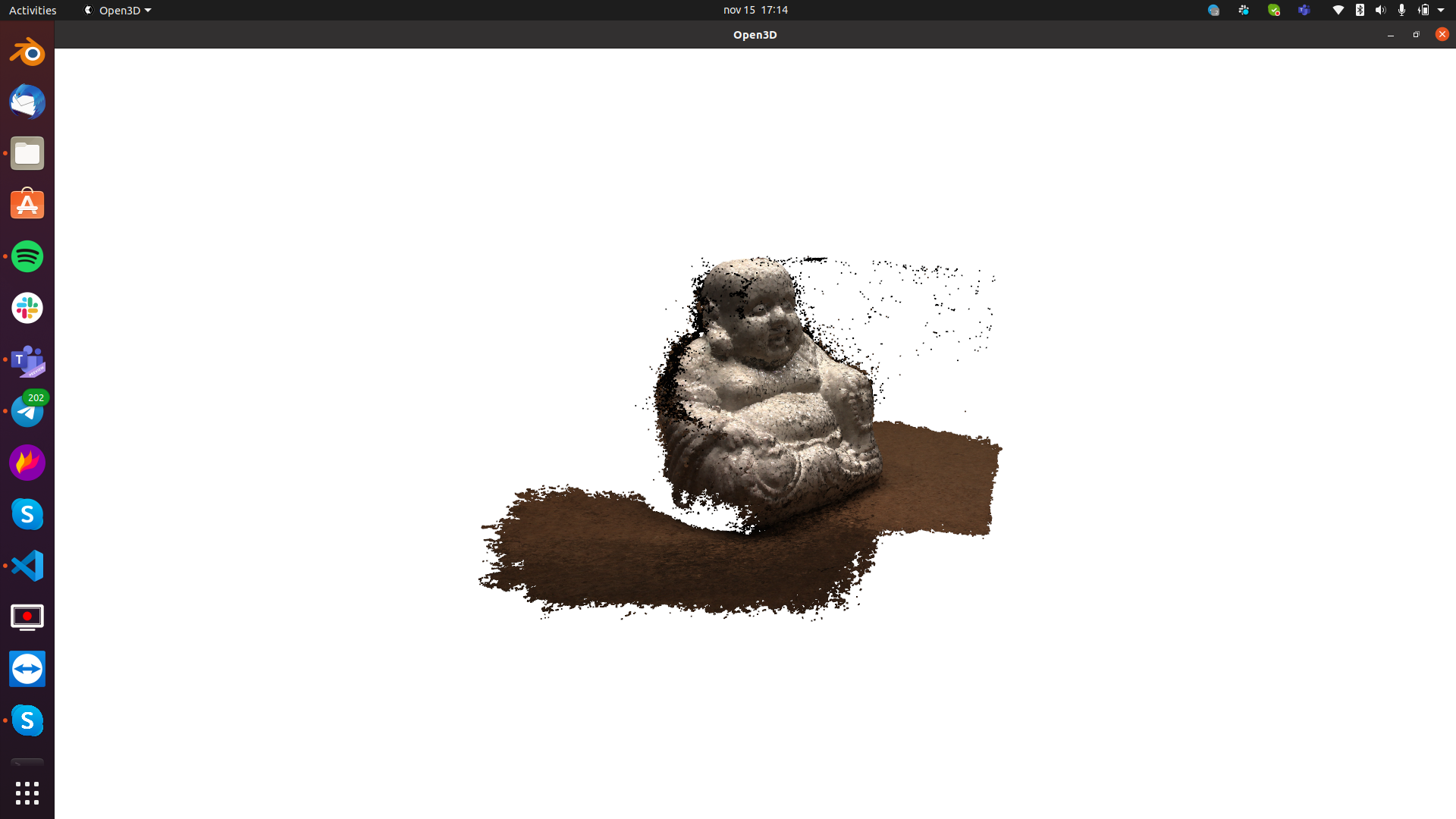} \\
         \includegraphics[height=0.24\linewidth]{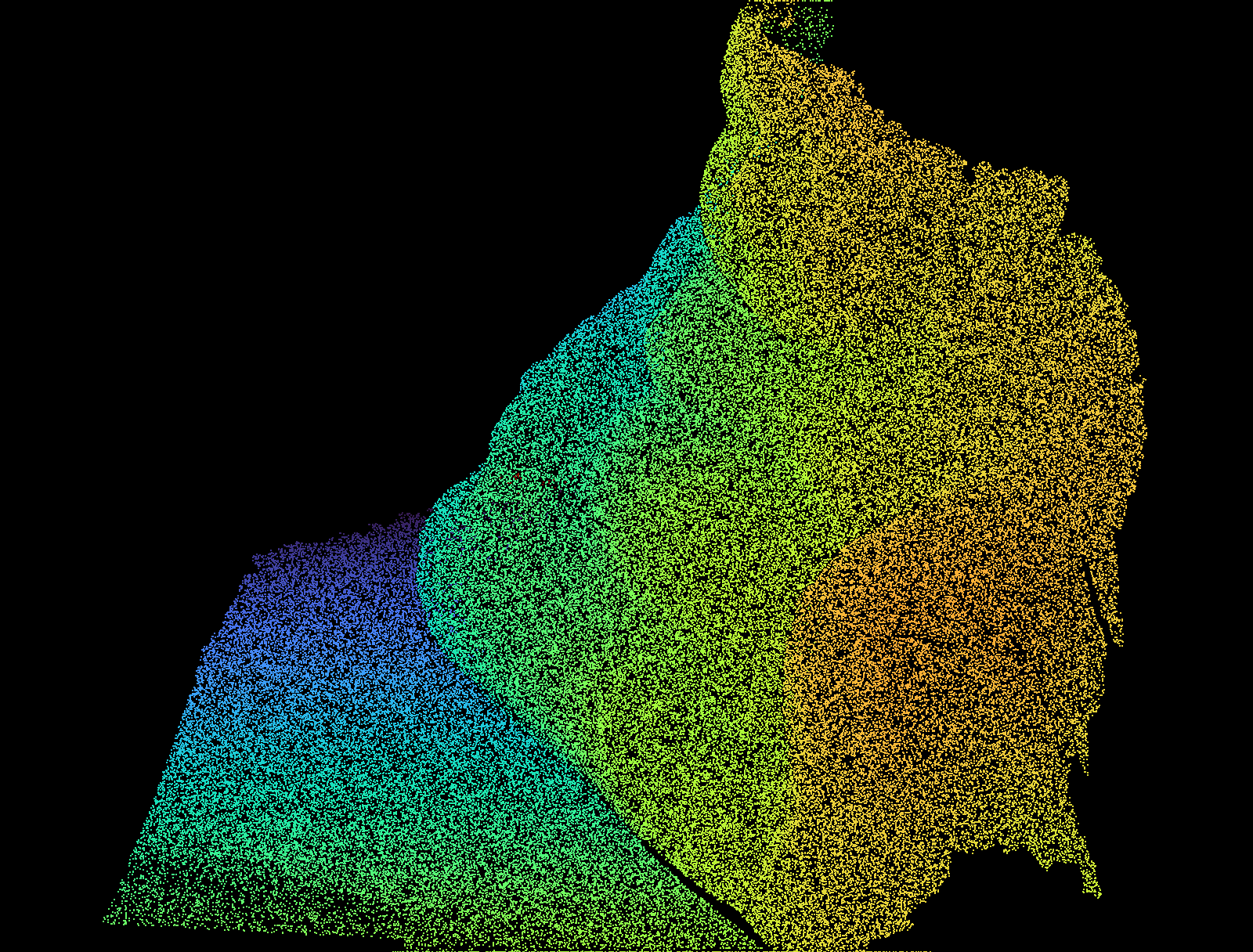} &
         \includegraphics[height=0.24\linewidth]{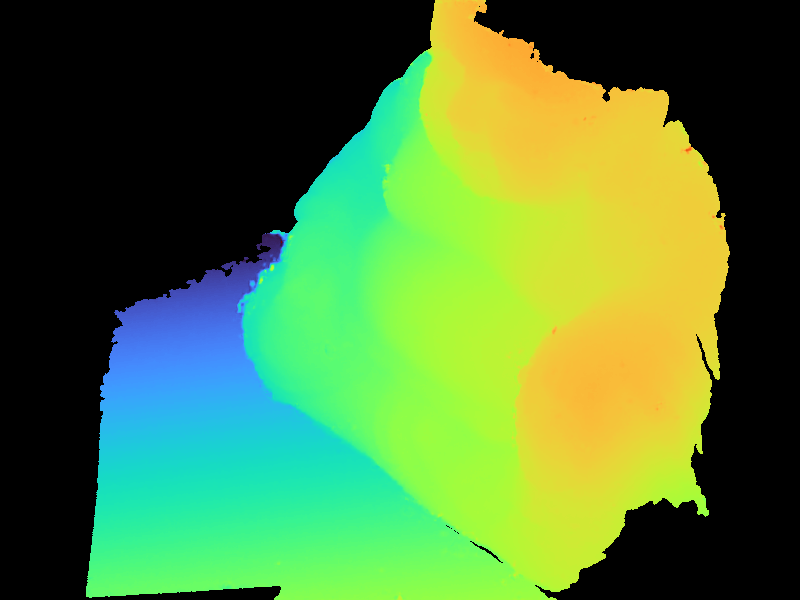} &
         \includegraphics[trim=22cm 12cm 15cm 10cm,clip,height=0.24\linewidth]{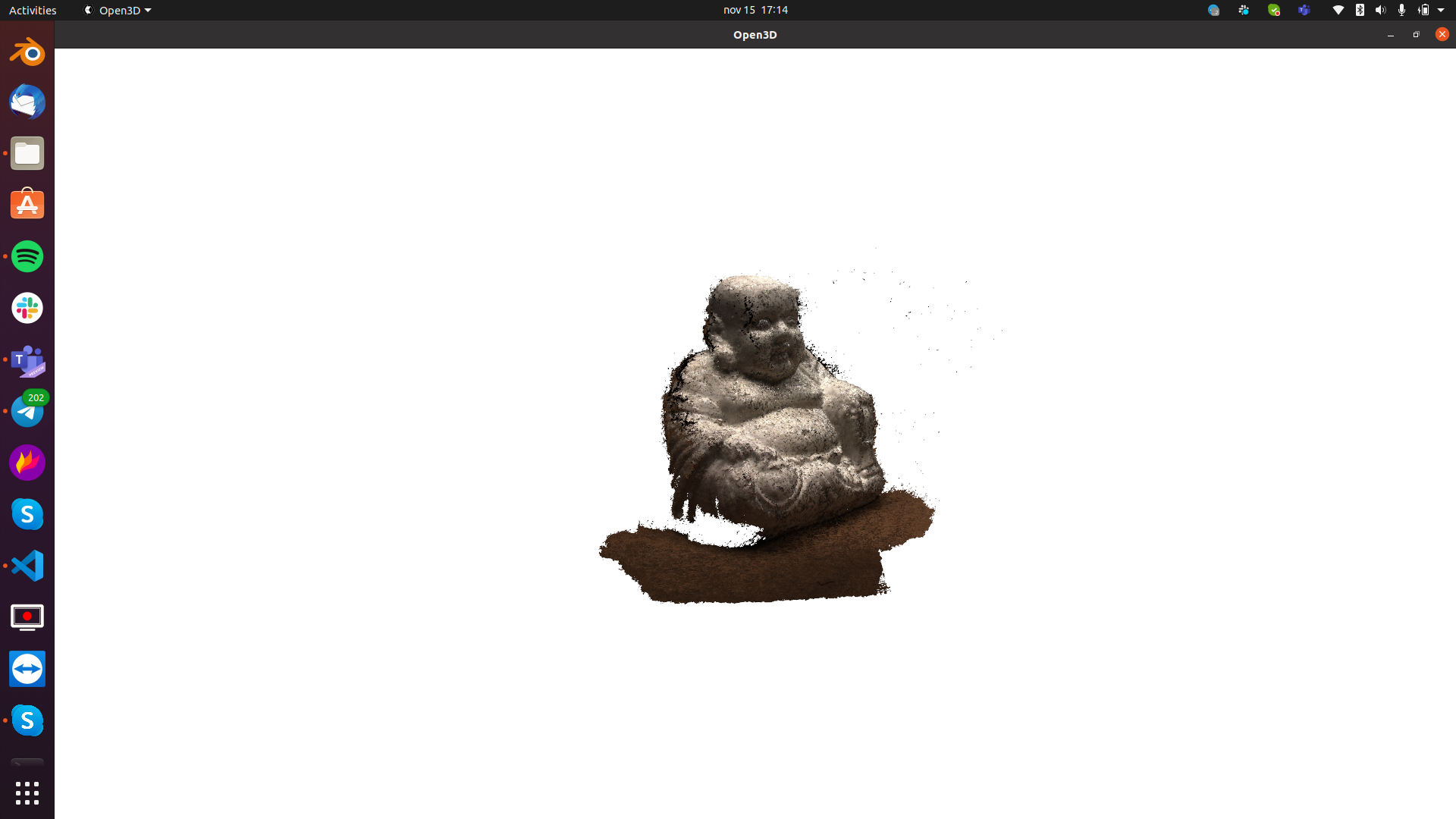} \\
    \end{tabular}
    }
    \caption{\textbf{Multi-View Guided Multi-View Stereo in action.} Deep MVS networks struggle at generalizing from synthetic to real images, yielding inaccurate depth maps and poor 3D reconstructions (top). By guiding the network with a set of sparse depth measurements, aggregated over the multiple views, we can greatly ameliorate the results (bottom). Depth maps are encoded with \url{turbo_r} colormap, while sparse depth hints on bottom row are densified by a $2\times2$ dilation filter to ease visualization in this teaser.}
    \label{fig:teaser}
\end{figure}

We argue that most of the challenges mentioned so far are inherent to the image domain itself. Thus, their impact could be significantly softened given the availability of additional information with different modalities, for instance, by having access to a sparse set of depth measurements perceived by an active sensor. Nowadays, such sensors are at hand and readily available as standalone off-the-shelf devices. Moreover, they are always more frequently integrated into consumer products like mobile phones and tablets (e.g., Apple iPhones and iPads). However, despite their ever-increasing diffusion, they often provide only sparse depth data (i.e., at a much lower resolution compared to standard cameras). 

The recent literature supports our intuition, highlighting the evidence of approaches effectively exploiting the synergy of color images with sparse depth data. For instance, in the case of depth completion \cite{Uhrig2017THREEDV}, fusion with stereo algorithms \cite{park2018high,park2019high} and networks \cite{Poggi_2019_CVPR,Cheng_2019_CVPR,wang20193d} or, more recently, with optical flow deep architectures \cite{Poggi_2021_ICCV} as well.

Driven by these facts, we propose a framework for guided multi-view stereo depth estimation. Assuming the availability of a sparse set of depth measurements acquired together with images, we modulate \cite{Poggi_2019_CVPR} the cost volume built by any state-of-the-art MVS network \cite{yao2018mvsnet,gu2020cascade,yan2020dense,cheng2020deep,wang2021patchmatchnet} to provide stronger guidance to the architecture towards inferring more accurate depth maps.
Moreover, by exploiting the possibility of having multiple sets of sparse depth points acquired from the different viewpoints of the source images, we introduce an integration mechanism to accumulate the multiple depth hints enabling modulating the cost volume inside the deep network with a higher density of guiding points. This allows to boost the performance of a MVS network, allowing it to infer more accurate depth maps, and consequently higher quality 3D reconstructions, for instance when trained on synthetic data and tested on real images, as shown in Fig. \ref{fig:teaser}.
To validate this claim, we run an exhaustive set of experiments by training a variety of state-of-the-art MVS architectures and their guided counterparts on the BlendedMVG \cite{yao2020blendedmvs} and DTU \cite{jensen2014large} datasets and assessing their accuracy on them. This proves that our framework consistently boosts the accuracy achievable with any considered deep networks in terms of depth map estimation and overall 3D reconstruction when guidance is available.
Our contributions are:

\begin{itemize}
    \item We propose the Guided Multi-View Stereo framework (gMVS), extending \cite{Poggi_2019_CVPR} to cope with our purposes. 
    Then, on top of that, we propose the Multi-View Guided Multi-View Stereo (mvgMVS) to exploit multiple sets of depth hints acquired from different viewpoints of the multi-view reconstruction task.
    
    \item We introduce coarse-to-fine guidance by applying cost volume modulation multiple times during the forward pass, compliantly to the coarse-to-fine strategy followed by recent MVS networks \cite{gu2020cascade,cheng2020deep,wang2021patchmatchnet}.
    
    \item We implement the proposed mvgMVS framework within five state-of-the-art deep architectures \cite{yao2018mvsnet,gu2020cascade,yan2020dense,cheng2020deep,wang2021patchmatchnet}, each one characterized by different regularization and optimization strategies.
    
\end{itemize}

\section{Related Work}

We review the literature relevant to our work concerning stereo vision, traditional MVS approaches, deep MVS networks and guided/conditioned deep learning frameworks.

\textbf{Stereo Matching.}
Predicting depth from a set of calibrated images is a fundamental task in computer vision and stereo matching \cite{Taxonomy_Stereo} represents the simplest approach for this purpose, leveraging two rectified images. This task has been faced through hand-crafted algorithms for years \cite{CostAggregationMethods,EnergyMinimization,SemiGlobalMatching}, until deep learning diffusion. At first, hand-crafted features used to compute matching costs were replaced with learned ones \cite{MC-CNN}, then end-to-end architectures \cite{DispNet, GCNet, GANet} became dominant on the stage \cite{SURVEY_STEREO_DEEP}.

\textbf{Multi-View Stereo.}
MVS extends stereo matching to an arbitrary number of images, acquired from known viewpoints.
Pre-deep learning techniques belongs to four categories, respectively reasoning about voxels \cite{MVSGraphCuts, SemantincMVS, MVSVoxelColoring}, surface evolution \cite{ AQuasiDenseSurfaceReconstruction}, matching patches \cite{furukawa2009accurate} or estimated depth maps \cite{Galliani_2015_ICCV}. 
The latter strategy results the most practical and efficient and has been embraced by modern deep learning MVS architectures.
The first proposal in this direction was MVSNet \cite{yao2018mvsnet}, a deep network building a variance-based cost volume through plane-sweep, then processed through 3D convolutions.
However, 3D CNNs are time and memory consuming and two main strategy have been propose to soften these constraints. The first consist of replacing 3D convolutions with 2D GRU unit \cite{yao2019recurrent,yan2020dense,wei2021aa}. 
The second implements multi-stage architectures capable of coarse-to-fine inference \cite{gu2020cascade,cheng2020deep,wang2021patchmatchnet} or pyramidal cost volumes \cite{yang2020cost}. 

\textbf{Depth Completion and Guided Frameworks.}
Two other research trends are relevant to our work. One concerns depth prediction from a single RGB image and sparse depth points, namely \textit{depth completion} \cite{uhrig2017sparsity,SparseToDense,GuidedNet,PENet}.
The second concerns the idea of conditioning deep features, by either acting in latent \cite{huang2017arbitrary,courville2017modulating,park2019semantic} or geometric space \cite{Cheng_2019_CVPR,Poggi_2019_CVPR,Poggi_2021_ICCV} via normalization or modulation. 

Our proposal to leverage sparse depth data within MVS networks takes inspiration from previous successes in stereo \cite{Poggi_2019_CVPR} and optical flow \cite{Poggi_2021_ICCV}. Nonetheless, we arguably extend the previous guided deep learning formulation under different aspects for our purposes. Specifically: i) considering multiple sources of depth hints, placed at different viewpoints, to increase the guide density and effectiveness and ii) applying modulation within coarse-to-fine architectures -- unexplored in previous works \cite{Poggi_2019_CVPR,Poggi_2021_ICCV}.

\section{Proposed framework}

In this section, we introduce our Multi-View Guided Multi-View Stereo (mvgMVS) framework. 
First, we review the background relevant to our proposal, specifically concerning deep MVS networks. Then, we cast the guided stereo matching framework \cite{Poggi_2019_CVPR} into the MVS setting and, finally, we extend it to deal with multi-view depth hints and coarse-to-fine architectures.

\begin{figure*}[t]
    \centering
    \renewcommand{\tabcolsep}{1pt}
     \includegraphics[trim=0cm 10cm 5cm 0cm,clip,width=0.7\textwidth]{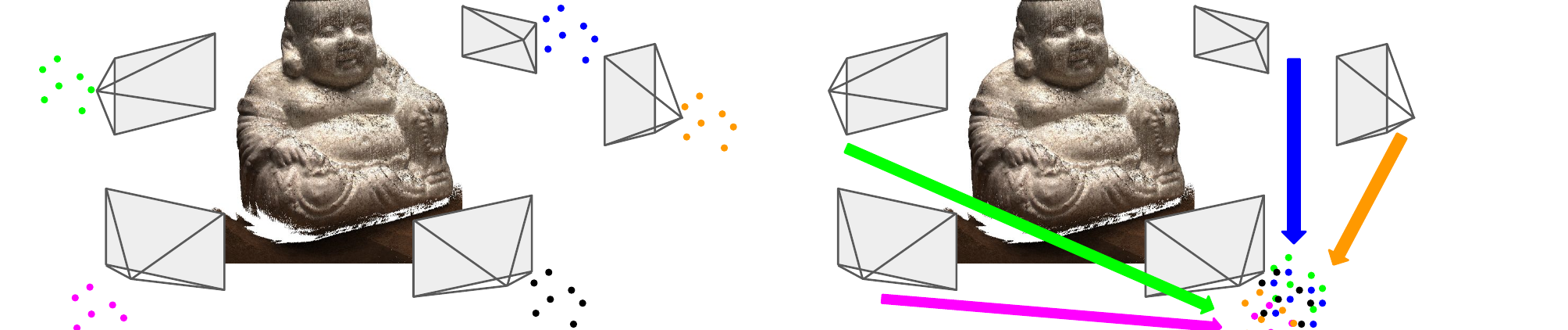}
    \caption{\textbf{Hints aggregation.} Depth hints from many views (left) can be aggregated on the reference image viewpoint (right).}
    \label{fig:hints_aggregation}
\end{figure*}

\subsection{Deep Multi-View Stereo background}
\label{sec:background}

Most learning-based MVS pipelines follow the same pattern. Given a set of $N$ images, one assumed as the reference and the other $N-1$ as source images, deep MVS networks process them to predict a global dense depth map aligned with the reference one.
To this aim, common to most deep networks designed for this purpose is the definition of a cost volume, encoding features similarity between pixels in the reference image and potential matching candidates from the source images. The latter are retrieved along the epipolar lines in the source views, given intrinsic and extrinsic parameters $K,E$ for any camera collecting the $N$ images involved.
Specifically, for a particular depth hypothesis $z \in [z_\text{min},z_\text{max}]$, features $\mathcal{F}_i$ extracted from a given source view $i$ are projected by means of an homography-based warping operation $\pi$.

\begin{equation}\label{eq:homog}
    \mathcal{F}^z_i = \pi(\mathcal{F}_i,z,K_0,E_0,K_i,E_i)
\end{equation}
Then, to encode the similarity between reference features $\mathcal{F}_0$ and $\mathcal{F}^z_i$, a variance-based volume is defined as follows

\begin{equation}
    \mathcal{V}(z) = \frac{\sum_{i=0}^N (\mathcal{F}_i^z - \mu)^2}{N}, \quad\quad \mu = \frac{\sum_{i=0}^N \mathcal{F}_i^z}{N}
\end{equation}
with $\mathcal{F}_i^z$ consisting of $\mathcal{F}_0$ for $i=0$.
Accordingly, for a given pixel, the lower the variance score, the more similar the features retrieved from the source views are and, thus, the more likely hypothesis $z$ is the correct depth for it.

However, implementing this solution requires high memory and results computationally complex. Consequently, several state-of-the-art networks \cite{gu2020cascade,yan2020dense,cheng2020deep,wang2021patchmatchnet} implement a coarse-to-fine solution.  
Specifically, a set of variance-based cost volumes are built as 

\begin{equation}
    \mathcal{V}_s(z) = \frac{\sum_{i=0}^N (\hat{\mathcal{F}}_{(i,s)}^d - \mu)^2}{N}, \quad\quad \mu = \frac{\sum_{i=0}^N \hat{\mathcal{F}}_{(i,s)}^z}{N}
\end{equation}
being $s$ a specific resolution or scale at which the cost volume is computed and $\hat{\mathcal{F}}^z_{(i,s)}$ features from image $i$ at resolution $s$ sampled as 

\begin{equation}\label{eq:homog_stage}
        \hat{\mathcal{F}}^z_{(i,s)} = \pi(\mathcal{F}_{(i,s)},z_s,K_{(0,s)},E_0,K_{(i,s)},E_i) 
\end{equation}
with $K_{(i,s)}$ being the intrinsic parameters for camera $i$ adjusted to resolution $s$ and $z_s$ sampled in a range $[z^s_\text{min},z^s_\text{max}]$ that differs at any scale.

\subsection{Guided Multi-View Stereo}
\label{sec:guided}

By assuming a setup made of a standard camera and a low-resolution depth sensor, for instance a LiDAR, we leverage the output of the latter to shape the behavior of a deep network estimating depth from a set of color images. When this set is limited to a single frame, a neural network is usually trained to \textit{complete} the sparse depth points \cite{Uhrig2017THREEDV} guided by the color image \cite{tang2020learning}. When multiple images are available, the mechanism often reverses, and depth measurements are used as \textit{hints} to guide the image-based estimation process. This strategy is implemented, for instance, by the Guided Stereo framework \cite{Poggi_2019_CVPR} applied to binocular stereo, by applying a Gaussian modulation to the features volume to peak it in correspondence of a depth hint $z$.

In analogy, this mechanism can be applied also to multi-view stereo, implementing a Guided Multi-View Stereo pipeline (gMVS). Indeed, the variance volume introduced in Sec. \ref{sec:background} can be conveniently modulated as well. In this case, since low variance encodes a high likelihood of the corresponding depth hypothesis $z$ to be correct, we flip the Gaussian curve to force the variance-based cost volume to have a minimum near depth hint $z^*$ 

\begin{equation}
\mathcal{V}'(z) = \left[1-v + v \cdot k \cdot \left(1 - e^{-\frac{(z-z^*)^2}{2c^2}}\right)\right] \cdot \mathcal{V}(z)
\label{eq:modulation}
\end{equation}
with $v$ being a binary mask equal to 1 for pixels with a valid hint (0 otherwise) and k, c being the amplitude and width of the Gaussian itself.
The gMVS formulation outlined so far extends the Guided Stereo framework \cite{Poggi_2019_CVPR} to MVS. In the remainder, we will introduce two significant additional contributions conceived explicitly for the MVS setup and the models designed for it.

\begin{figure}
    \centering
    \renewcommand{\tabcolsep}{1pt}
     \includegraphics[trim=0cm 6cm 4cm 0cm,clip,width=0.4\textwidth]{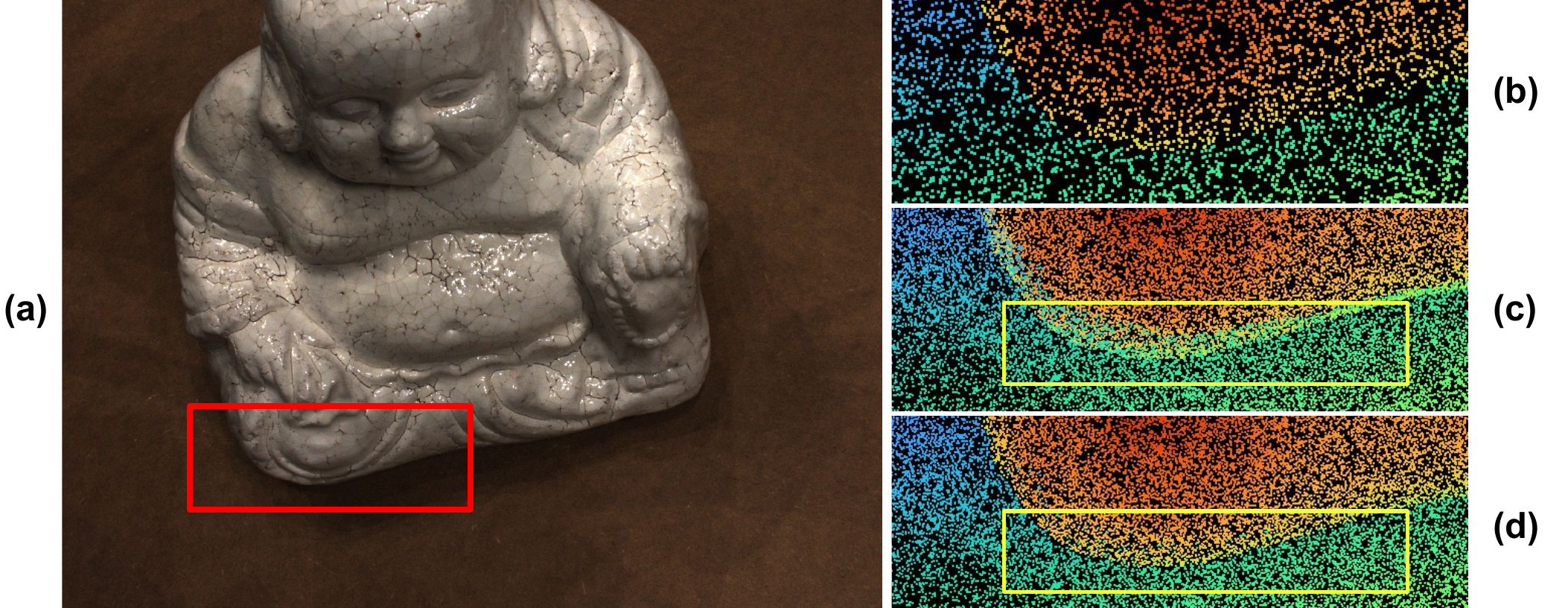}
         
    \caption{\textbf{Depth hints filtering.} On left, reference image (a). On right, sparse hints over the region inside red rectangle in (a), respectively from the single viewpoint (b), aggregated over multiple viewpoints (c) and filtered (d). Regions in yellow rectangles in (c) and (d) highlight the effect of filtering.
    Depth points are densified to ease visualization.}
    \label{fig:hints_filtering}
\end{figure}

\subsection{Multi-View Guided Multi-View Stereo}

MVS relies on the availability of multiple images acquired from different viewpoints. Moreover, we assume the availability of sparse depth measurements registered with the colour images in our setup. Therefore, a different set of hints is available for each source image. In such a case, we argue that aggregating the multiple sets of depth hints from each viewpoint can provide stronger guidance to the network and further improve the results of the baseline gMVS framework. To this aim, we perform two main steps.

\textbf{Depth hints aggregation.}
Given a pixel having homogeneous 2D coordinates $q_i$ from any source image $i \in [1,N]$ for which a depth value $d^*_{q_i}$ is available, the 3D coordinates $p_0$ in the reference image viewpoint are obtained as:
\begin{equation}
    p_0 = E_0 E_i^{-1} p_i \quad\quad\text{{with}} \quad\quad p_i = d^*_{q_i} K_i^{-1}q_i
\end{equation}
From $p_0$, we can get the new depth hint $d^*_{q_0}$ expressed in the reference image viewpoint, and project it on the image plane according to $K_0$ at coordinates $q_0$.

This allows to aggregate depth hints on the reference view, as shown in Fig. \ref{fig:hints_aggregation}, and thus obtain a denser depth hints map to modulate the volume in the network with stronger guidance. We refer to this extension of the gMVS framework as Multi-View Guided Multi-View Stereo (mvgMVS)

\textbf{Depth hints filtering.}
Because of the different viewpoints, some of the depth measurements acquired in one of the source views may belong to occluded regions in the reference view. However, given the sparse nature of the hints, this would cause the aggregation of several wrong values if we would limit to naively projecting them across the views without reasoning about their visibility, as shown in Fig. \ref{fig:hints_filtering} (b). As a consequence, we would guide the deep network with wrong depth hints, harming its accuracy.
To detect and remove these outliers, we deploy the filtering strategy by Zhao et al. \cite{ZhaoYiming_IEEE_2021}, defining as outlier any pixels $q_0$ for which exists at least a pixel $s$ in its neighbourhood $S(q_0)$ such that:

\begin{itemize}
    \item $q_0$ changes the relative position with respect to $s$, because occluded. This occurs if the difference between $q_0$ and $s$ pixels coordinates and angles (in spherical coordinates) have different sign, i.e. if either $(x_{q_0} - x_{s})(\theta_{q_0}-\theta_{s})$ or $(y_{q_0} - y_{s})(\phi_{q_0}-\phi_{s})$ are negative
    
    \item $q_0$ distance from the camera is much higher compared to $s$, i.e. $d_{q_0} > d_{s} + \varepsilon$, with $\varepsilon$ set according to the specific dataset used
\end{itemize}
Although simple, this strategy allows for removing most of the outliers at a minor computational cost, as shown in Fig. \ref{fig:hints_filtering} (c). We will show in our ablation experiments how this step is necessary to achieve optimal guidance. We refer to this final implementation as filtered mvgMVS (fmvgMVS).

\subsection{Coarse-to-Fine Guidance}
\label{sec:coarsetofine}

Unlike deep stereo networks, which usually build a single volume processed through stacked 3D convolutions, MVS networks are often designed to embody coarse-to-fine estimation to reduce the computational burden, as introduced previously in Sec. \ref{sec:background}. We argue that any of the multiple cost volumes built by the network represent a possible entry point for guiding the network. Accordingly, we then modulate any $\mathcal{V}_s$ during the forward pass

\begin{equation}
\mathcal{V}_s'(z_s) = \left[1-v_s + v_s \cdot k \cdot \left(1 - e^{-\frac{(z_s-z^*_s)^2}{2c^2}}\right)\right] \cdot \mathcal{V}_s(z_s)
\end{equation}
with $v_s$ and $z^*_s$ being respectively the binary mask $v$ and the depth hints map $z^*$ downsampled to resolution $s$, with nearest-neighbor interpolation.
Our experiments will show how the stronger guidance yielded by these multiple modulations improves the overall network accuracy.

\section{Experimental results}

In this section, we collect the outcome of our experiments describing, at first, the datasets involved in our evaluation, details concerning the framework implementation, the networks evaluated and the training protocol. Source code is available at \url{https://github.com/andreaconti/multi-view-guided-multi-view-stereo}.

\subsection{Datasets}

We begin by introducing the datasets involved in our experiments. Since none of the existing MVS data collection provides sparse depth points, we simulate the availability of sparse hints by randomly sampling them from ground-truth depth maps, similarly to \cite{Poggi_2019_CVPR,Poggi_2021_ICCV}. Consequently, for our experiments, we can select only datasets providing such information, i.e. we cannot evaluate on Tank \& Temples \cite{knapitsch2017tanks}.

\textbf{BlendedMVG.} This dataset \cite{yao2020blendedmvs} collects about 110K images sampled from about 500 scenes. It has been created by applying a 3D reconstruction pipeline to recover high-quality textured meshes from images of well-selected scenes. Then, meshes are rendered to color images and depth maps. Following \cite{yao2020blendedmvs} we retain 8 sequences for validation and 7 for testing, using the rest for training each network involved in our experiments. 

\textbf{DTU \cite{aanaes2016large}.} This indoor dataset counts 124 different scenes, all sharing the very same camera trajectory. Images are acquired with a structured light scanner mounted on a robot arm, using one of the cameras in the scanner itself.
We select training, validation and testing splits following existing works \cite{yao2018mvsnet,gu2020cascade,yan2020dense,cheng2020deep,wang2021patchmatchnet}. In particular, we evaluate on the testing split both networks trained on BlendedMVG alone or after being fine-tuned on the DTU training set.

\subsection{Implementation details}

Our framework is implemented in PyTorch, starting from existing solutions \cite{wang2021patchmatchnet}. Concerning gMVS, we simulate the availability of sparse depth hints by randomly sampling 3\% of pixels from the ground-truth depth maps.
Following \cite{Poggi_2019_CVPR,Poggi_2021_ICCV}, we set $k=10$ and $c=0.01$. Regarding filtering, we set $\varepsilon=3$.
We conduct our experiments implementing gMVS and variants with five state-of-the-art networks.

\textbf{MVSNet \cite{yao2018mvsnet}.} The very first deep network for MVS: it builds a single variance volume and process it through 3D convolutions -- similarly to 3D stereo networks \cite{Kendall_2017_ICCV} -- and estimates depth at a quarter of the input resolution.

\textbf{D$^2$HC-RMVSNet \cite{yan2020dense}.} A recurrent architecture, replacing 3D convolutions with 2D convolutional LSTM to reduce memory requirements.

\textbf{CAS-MVSNet \cite{gu2020cascade}.} It implements a cascade cost volume formulation, inferring depth in a coarse-to-fine manner to achieve higher efficiency.

\textbf{UCSNet \cite{cheng2020deep}.} It builds Adaptive Thin Volumes for coarse-to-fine processing. The volumes consist of only a few depth hypotheses selected by modeling uncertainty.

\textbf{PatchMatchNet \cite{wang2021patchmatchnet}.} A very efficient model, implementing a differentiable variant of the PatchMatch algorithm \cite{barnes2009patchmatch} within a deep network.

Any network is implemented by integrating the authors' code in our framework and following their default configuration -- except for the number of depth hypotheses used by MVSNet and D$^2$HC-RMVSNet, set to 128 because of memory constraints. 
During both training and evaluation, if the final output of the original network is lower than the input resolution, it is upsampled to the original size through interpolation.

\subsection{Training and testing protocol}

We set the number of images processed by the networks to 5, both during training and testing. Accordingly, we accumulate depth hints coming from 5 views for mvgMVS.

\textbf{Training schedule.} We train each network for 100K iterations on the BlendedMVG dataset on $576\times768$ images, with a constant learning rate of 10$^{-3}$ -- except D$^2$HC-RMVSNet, for which it was set to 10$^{-4}$ to avoid instability. Any training has been carried out on a single Titan Xp GPU, allowing only for a single sample per batch -- except for PatchMatchNet, for which batch 2 fits in memory.

We also fine-tune each network for 50K further iterations on the DTU training set, processing $512\times640$ images and using the hyper-parameters as detailed for BlendedMVG. 

\textbf{Testing protocol.} We test the networks on the BlendedMVG testing sequences and on the DTU testing split. For each dataset, we report the percentage of pixels in the estimated depth map having an error larger than $\tau$ -- respectively in pixels and millimetres on the two datasets, with thresholds set to 1, 2, 3 and 4.
Concerning DTU, we also evaluate the quality of reconstructed point clouds: in the former case, we report \textit{accuracy} and \textit{completeness} metrics defined as in \cite{aanaes2016large} and their average -- {the lower the better}. Fused point clouds are obtained as in \cite{wang2021patchmatchnet}.

\subsection{Ablation study}

We start by studying the impact of the different components in our framework, with the main emphasis on mvgMVS extension and coarse-to-fine modulation.

\textbf{Multi-View Guided MVS.} We first measure the improvements yielded by multi-view guidance. To this aim, we run experiments with MVSNet, by training different variants on the BlendedMVG training split and evaluating on the testing sequences.
Tab. \ref{tab:ablation} collects the outcome of this experiment. From top to bottom, we report the error rates achieved by the original MVSNet architecture, by a variant implementing the baseline guided MVS framework described in Sec. \ref{sec:guided} (-g), followed by mvgMVS versions respectively without (-mvg) and with (-fmvg) filtering.

\begin{table}[]
\centering
\resizebox{0.4\textwidth}{!}{%
\begin{tabular}{l|r|cccc}
\specialrule{.2em}{.1em}{.1em}
\multicolumn{1}{l|}{Network} & Hints dens. &  \multicolumn{1}{l}{\textgreater{}1 Px. } & \multicolumn{1}{l}{\textgreater{}2 Px. } & \multicolumn{1}{l}{\textgreater{}3 Px. } & \multicolumn{1}{l}{\textgreater{}4 Px. } \\ 
\midrule
MVSNet \cite{yao2018mvsnet} & - & 0.139 & 0.073 & 0.046 & 0.031 \\
\midrule
MVSNet-g & 0.03 & 0.095 & 0.046 & 0.027 & 0.018 \\
MVSNet-mvg & 0.03 & 0.081 & 0.040 & 0.024 & 0.016 \\
MVSNet-fmvg & 0.03 & \bfseries 0.076 & \bfseries 0.037 & \bfseries 0.023 & \bfseries 0.015 \\
\midrule
MVSNet-g & 0.15 & 0.068 & 0.032 & 0.020 & 0.013 \\
\specialrule{.2em}{.1em}{.1em}
\end{tabular}%
}
  
\centering
\caption{\textbf{Ablation study -- guiding strategy.} Results on BlendedMVG \cite{yao2020blendedmvs} testing scans.}
\label{tab:ablation}
  
\end{table}

\begin{table}[]
\centering
\resizebox{0.4\textwidth}{!}{%
\begin{tabular}{l|r|cccc}
\specialrule{.2em}{.1em}{.1em}
\multicolumn{1}{l|}{Network} & Hints dens. &  \multicolumn{1}{l}{\textgreater{}1 Px. } & \multicolumn{1}{l}{\textgreater{}2 Px. } & \multicolumn{1}{l}{\textgreater{}3 Px. } & \multicolumn{1}{l}{\textgreater{}4 Px. } \\ 
\midrule
MVSNet-fmvg & 0.03 & \bfseries 0.076 & \bfseries 0.037 & \bfseries 0.023 & \bfseries 0.015 \\
\midrule
GuideNet-fmvg & 0.03 & 0.290 & 0.124 & 0.080 & 0.058 \\ 
\specialrule{.2em}{.1em}{.1em}
\end{tabular}%
}
  
\centering
\caption{\textbf{Ablation study -- mvgMVS versus depth completion.} Results on BlendedMVG \cite{yao2020blendedmvs} testing scans.}
\label{tab:completion}
  
\end{table}

\begin{table}[t]
\centering
\resizebox{0.4\textwidth}{!}{%
\begin{tabular}{l|r|cccc}
\specialrule{.2em}{.1em}{.1em}
\multicolumn{1}{l|}{Network} & Stages &  \multicolumn{1}{l}{\textgreater{}1 Px. } & \multicolumn{1}{l}{\textgreater{}2 Px. } & \multicolumn{1}{l}{\textgreater{}3 Px. } & \multicolumn{1}{l}{\textgreater{}4 Px. } \\ 
\midrule
CAS-MVSNet \cite{gu2020cascade} & - & 0.071 & 0.036 & 0.023 & 0.016 \\
\midrule
CAS-MVSNet-fmvg & 1 & 0.057 & 0.024 & 0.014 & 0.010 \\
CAS-MVSNet-fmvg & 2 & 0.084 & 0.042 & 0.027 & 0.019 \\
CAS-MVSNet-fmvg & 3 & 0.078 & 0.041 & 0.027 & 0.020 \\
CAS-MVSNet-fmvg & All & \bfseries 0.048 & \bfseries 0.018 & \bfseries 0.012 & \bfseries 0.009 \\
\specialrule{.2em}{.1em}{.1em}
\end{tabular}%
}
  
\centering
\caption{\textbf{Ablation study -- multi-stage guidance.} Results on BlendedMVG \cite{yao2020blendedmvs} testing scans.}
\label{tab:coarsetofine}
  
\end{table}

Starting from the gMVS baseline, it consistently achieves reduced error rates compared to MVSNet by exploiting the sparse depth guidance. Concerning mvgMVS, there are further improvements thanks to the aggregation of multiple sets of depth hints coming from the 5 different viewpoints. Nonetheless, even if this strategy increases the hints density from 3\% up to roughly 15\%, the improvement might appear not significant as one might expect with a more extensive set of hints. This fact is due to the several hints in non-visible parts of the source images that are wrongly projected in the reference point of view, as discussed previously. 
Indeed, by filtering out these outliers and consequently reducing the hints density to about 14\%, we can improve the performance of MVSNet further when guided by mvgMVS.
At the bottom of the table, we also report the performance achieved by MVSNet when guided by the baseline gMVS implementation and 15\% hints density. Not surprisingly, having a higher density of depth hints from the single reference viewpoint is more effective than aggregating them over multiple viewpoints because they are not affected by visibility and possible collisions between projected points. 
However, fmvgMVS achieves performance close to what attainable with a depth sensor providing a much denser guide.

To conclude this study, we also compare the performance of our MVSNet-fmvg with a depth completion framework. Purposely, we select GuideNet \cite{GuidedNet} and train it to process single, RGB images and multi-view aggregated sparse depth points -- the very same used to guide MVSNet -- for 100K iterations on BlendedMVG as done for MVSNet. Tab. \ref{tab:completion} directly compares the error rates achieved by both highlighting how, when multiple sets of depth hints are available, the guided multi-view framework yields better depth maps compared to a depth completion approach.

\begin{table}[t]
\centering
\resizebox{0.4\textwidth}{!}{%
\begin{tabular}{l|cccc}
\specialrule{.2em}{.1em}{.1em}
\multicolumn{1}{l|}{Network} & \multicolumn{1}{l}{\textgreater{}1 Px.} & \multicolumn{1}{l}{\textgreater{}2 Px.} & \multicolumn{1}{l}{\textgreater{}3 Px.} & \multicolumn{1}{l}{\textgreater{}4 Px.} \\ 
\midrule
MVSNet \cite{yao2018mvsnet} & 0.139 & 0.073 & 0.046 & 0.031 \\
MVSNet-fmvg & \bfseries 0.076 & \bfseries 0.037 & \bfseries 0.023 & \bfseries 0.015 \\
\midrule
D$^2$HC-RMVSNet \cite{yan2020dense} & 0.174 & 0.094 & 0.059 & 0.040 \\
D$^2$HC-RMVSNet-fmvg & \bfseries 0.081 & \bfseries 0.041 & \bfseries 0.025 & \bfseries 0.017 \\
\midrule
CAS-MVSNet \cite{gu2020cascade} & 0.071 & 0.036 & 0.023 & 0.016 \\
CAS-MVSNet-fmvg & \bfseries 0.048 & \bfseries 0.018 & \bfseries 0.012 & \bfseries 0.009 \\
\midrule
UCSNet \cite{cheng2020deep} & 0.071 & 0.038 & 0.024 & 0.017 \\
UCSNet-fmvg & \bfseries 0.040 & \bfseries 0.018 & \bfseries 0.011 & \bfseries 0.008 \\
\midrule
PatchMatchNet \cite{wang2021patchmatchnet} & 0.075 & 0.039 & 0.025 & 0.018 \\
PatchMatchNet-fmvg & \bfseries 0.062 & \bfseries 0.033 & \bfseries 0.022 & \bfseries 0.016 \\
\specialrule{.2em}{.1em}{.1em}
\end{tabular}%
}
  
\centering
\caption{\textbf{Evaluation on BlendedMVG \cite{yao2020blendedmvs} testing scans.} Comparison between original MVS networks\cite{yao2018mvsnet,gu2020cascade,yan2020dense,cheng2020deep,wang2021patchmatchnet} and their guided counterparts.}
\label{tab:blended}
  
\end{table}

\begin{table*}[]
\centering
\resizebox{0.9\textwidth}{!}{%
\begin{tabular}{ccc}
\begin{tabular}{l}
\multicolumn{1}{c}{} \\
\specialrule{.2em}{.1em}{.1em}
\multicolumn{1}{l}{Network} 
\\ \midrule
MVSNet \cite{yao2018mvsnet}  \\
MVSNet-fmvg \\
\midrule
D$^2$HC-RMVSNet \cite{yan2020dense} \\
D$^2$HC-RMVSNet-fmvg \\
\midrule
CAS-MVSNet \cite{gu2020cascade} \\
CAS-MVSNet-fmvg  \\
\midrule
UCSNet \cite{cheng2020deep}  \\
UCSNet-fmvg \\
\midrule
PatchMatchNet \cite{wang2021patchmatchnet} \\
PatchMatchNet-fmvg  \\

\specialrule{.2em}{.1em}{.1em}
\end{tabular}%
& \quad

\begin{tabular}{cccc|ccc}
\multicolumn{4}{c}{Depth map evaluation} & \multicolumn{3}{c}{Point cloud evaluation} \\
\specialrule{.2em}{.1em}{.1em}
\multicolumn{1}{l}{\textgreater{}1 mm} & \multicolumn{1}{l}{\textgreater{}2 mm} & \multicolumn{1}{l}{\textgreater{}3 mm} & \multicolumn{1}{l}{\textgreater{}4 mm} 
& \multicolumn{1}{|l}{Acc. (mm)}
& \multicolumn{1}{l}{Comp. (mm)}
& \multicolumn{1}{l}{Avg. (mm)}
\\ \midrule
0.658 & 0.457 & 0.368 & 0.326 & 0.764 & 0.468 & 0.616 \\
\bfseries 0.393 & \bfseries 0.227 & \bfseries 0.194 & \bfseries 0.180 & \bfseries 0.383 & \bfseries 0.264 & \bfseries 0.324 \\
\midrule
0.708 & 0.519 & 0.423 & 0.372 & 0.764 & 0.586 & 0.675 \\
\bfseries 0.401 & \bfseries 0.177 & \bfseries 0.134 & \bfseries 0.115 & \bfseries 0.393 & \bfseries 0.234 & \bfseries 0.314 \\
\midrule
0.558 & 0.385 & 0.330 & 0.303 & 0.589 & 0.310 & 0.450 \\
\bfseries 0.323 & \bfseries 0.243 & \bfseries 0.220 & \bfseries 0.207 & \bfseries 0.345 & \bfseries 0.286 & \bfseries 0.316\\
\midrule
0.541 & 0.402 & 0.357 & 0.333 & 0.561 & 0.344 & 0.453 \\
\bfseries 0.199 & \bfseries 0.174 & \bfseries 0.164 & \bfseries 0.157 & \bfseries 0.290 & \bfseries 0.264 & \bfseries 0.277 \\
\midrule
0.627 & 0.440 & 0.370 & 0.335 & 0.574 & 0.484 & 0.529 \\
\bfseries 0.446 & \bfseries 0.328 & \bfseries 0.301 & \bfseries 0.287 & \bfseries 0.339 & \bfseries 0.297 & \bfseries 0.318 \\

\specialrule{.2em}{.1em}{.1em}
\end{tabular}%
& \quad

\begin{tabular}{cccc|ccc}
\multicolumn{4}{c}{Depth map evaluation} & \multicolumn{3}{c}{Point cloud evaluation} \\
\specialrule{.2em}{.1em}{.1em}
\multicolumn{1}{l}{\textgreater{}1 mm} & \multicolumn{1}{l}{\textgreater{}2 mm} & \multicolumn{1}{l}{\textgreater{}3 mm} & \multicolumn{1}{l}{\textgreater{}4 mm} 
& \multicolumn{1}{|l}{Acc. (mm)}
& \multicolumn{1}{l}{Comp. (mm)}
& \multicolumn{1}{l}{Avg. (mm)}
\\ \midrule
0.555 & 0.340 & 0.268 & 0.237 & 0.635 & 0.304 & 0.470 \\
\bfseries 0.219 & \bfseries 0.103 & \bfseries 0.081 & \bfseries 0.072 & \bfseries 0.324 & \bfseries 0.235 & \bfseries 0.280 \\
\midrule
0.630 & 0.423 & 0.329 & 0.283 & 0.662 & 0.342 & 0.502\\
\bfseries 0.168 & \bfseries 0.079 & \bfseries 0.061 & \bfseries 0.054 & \bfseries 0.327 & \bfseries 0.240 & \bfseries 0.284 \\
\midrule
0.480 & 0.307 & 0.257 & 0.233 & 0.528 & 0.262 & 0.395 \\
\bfseries 0.082 & \bfseries 0.056 & \bfseries 0.047 & \bfseries 0.042 & \bfseries 0.228 & \bfseries 0.279 & \bfseries 0.254 \\
\midrule
0.506 & 0.332 & 0.277 & 0.254 & 0.551 & 0.272 & 0.412 \\
\bfseries 0.119 & \bfseries 0.105 & \bfseries 0.098 & \bfseries 0.095 & \bfseries 0.319 & \bfseries 0.281 & \bfseries 0.300 \\
\midrule
0.475 & 0.310 & 0.260 & 0.236 & 0.461 & 0.298 & 0.380 \\
\bfseries 0.336 & \bfseries 0.228 & \bfseries 0.204 & \bfseries 0.193 &  \bfseries 0.325 & \bfseries 0.230 & \bfseries 0.278 \\

\specialrule{.2em}{.1em}{.1em}
\end{tabular}%
\\
 & \textbf{(a) trained on BlendedMVG} & \textbf{(b) fine-tuned on DTU} \\
\end{tabular}
}
  
\centering
\caption{\textbf{Evaluation on DTU \cite{aanaes2016large} testing scans.} Comparison between MVS networks \cite{yao2018mvsnet,gu2020cascade,yan2020dense,cheng2020deep,wang2021patchmatchnet} and guided counterparts, trained on BlendedMVG and tested (a) without re-train or (b) after fine-tuning on DTU training split.}
\label{tab:dtu}
  
\end{table*}

\begin{figure*}
    \centering
    \renewcommand{\tabcolsep}{1pt}
    \begin{tabular}{cccccc}
    \scriptsize \textit{RGB} &  \multicolumn{2}{c}{\scriptsize \textit{D$^2$HC-RMVSNet} \cite{yan2020dense}} & \scriptsize \textit{Hints} & \multicolumn{2}{c}{\scriptsize \textit{D$^2$HC-RMVSNet-fmvs}} \\
     \includegraphics[height=0.08\linewidth]{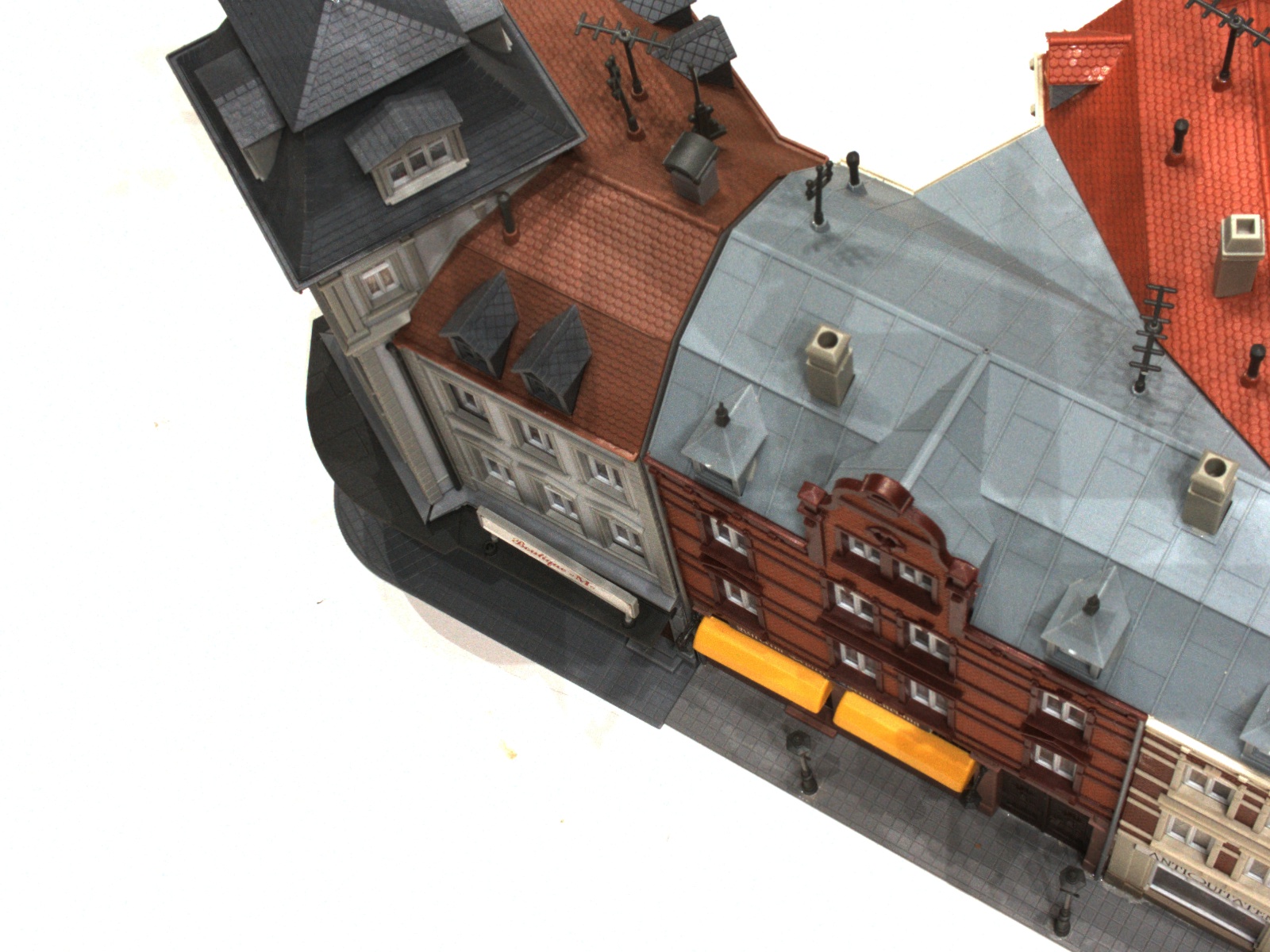} &
     \includegraphics[height=0.08\linewidth]{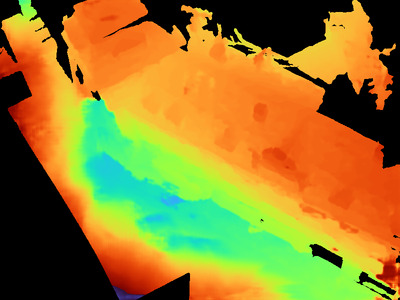} &
     \includegraphics[trim=10cm 10cm 11cm 4cm,clip,height=0.08\linewidth]{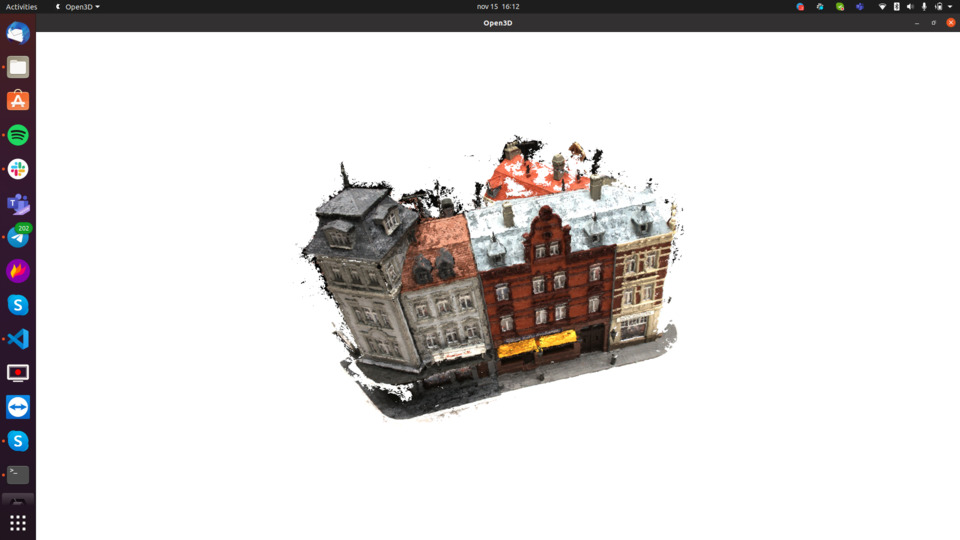} &
     \includegraphics[height=0.08\linewidth]{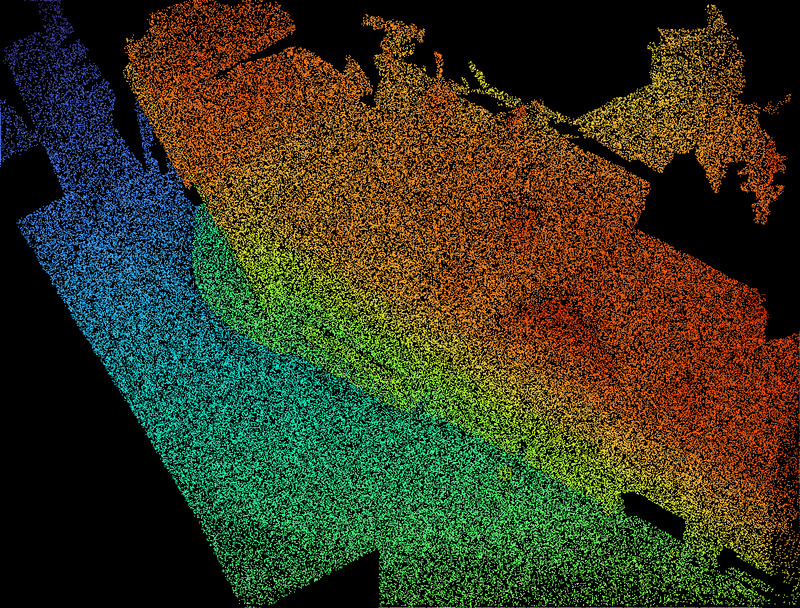} &
     \includegraphics[height=0.08\linewidth]{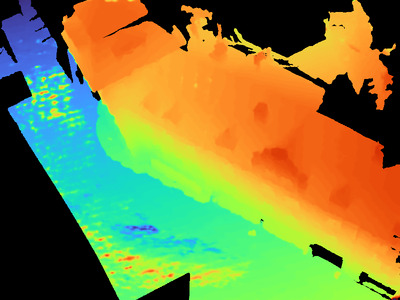} &
     \includegraphics[trim=10cm 10cm 11cm 4cm,clip,height=0.08\linewidth]{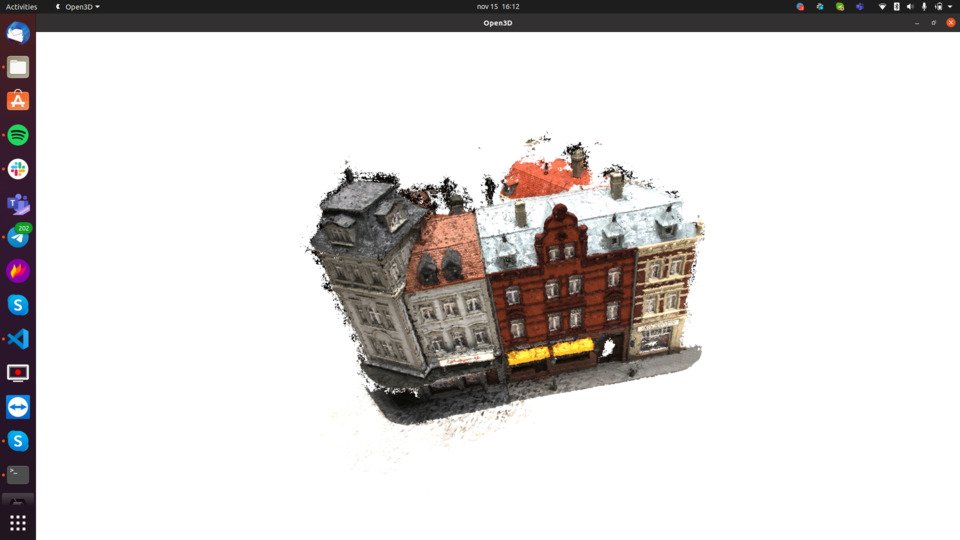} 
     \\
     \hline
     &  \multicolumn{2}{c}{\scriptsize \textit{CAS-MVSNet} \cite{gu2020cascade}} & &  \multicolumn{2}{c}{\scriptsize \textit{CAS-MVSNet-fmvs}} \\
     \includegraphics[height=0.08\linewidth]{images/qual-dtu/114-rgb.jpg} &
     \includegraphics[height=0.08\linewidth]{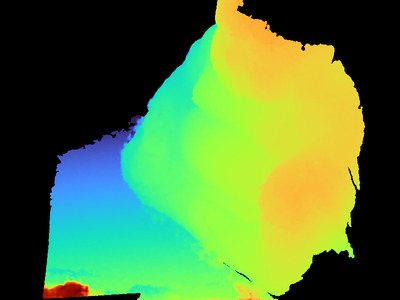} &
     \includegraphics[trim=11cm 6cm 7.5cm 5cm,clip,height=0.08\linewidth]{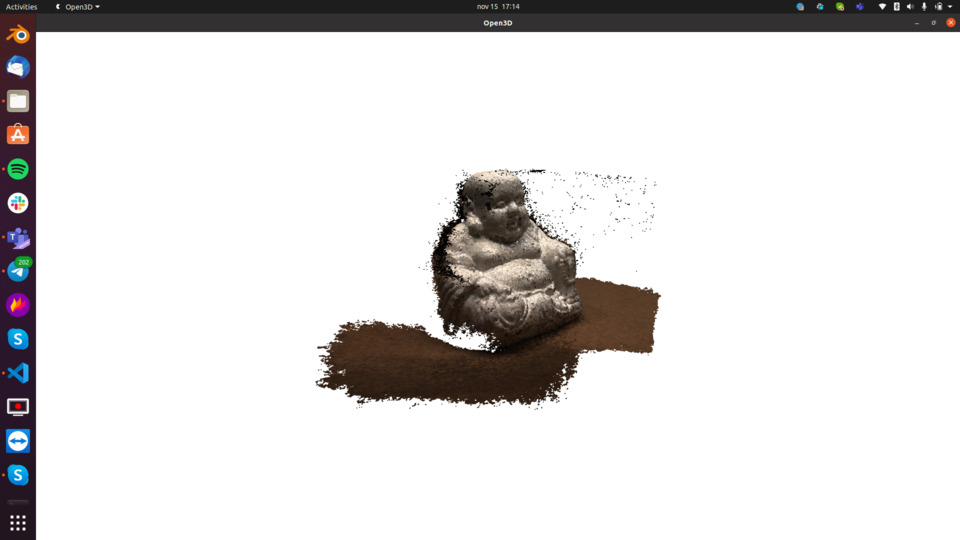} &
     \includegraphics[height=0.08\linewidth]{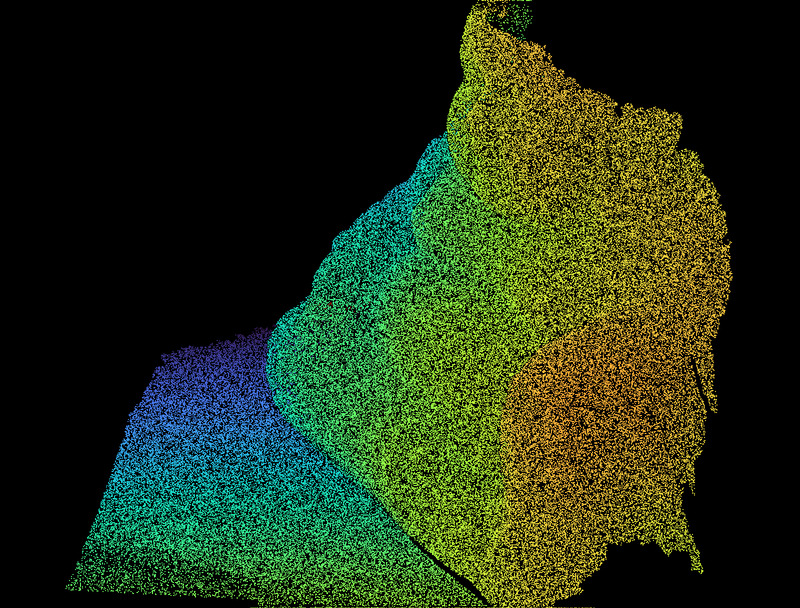} &
     \includegraphics[height=0.08\linewidth]{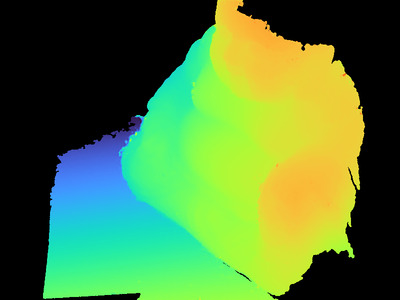} &
     \includegraphics[trim=11cm 6cm 7.5cm 5cm,clip,height=0.08\linewidth]{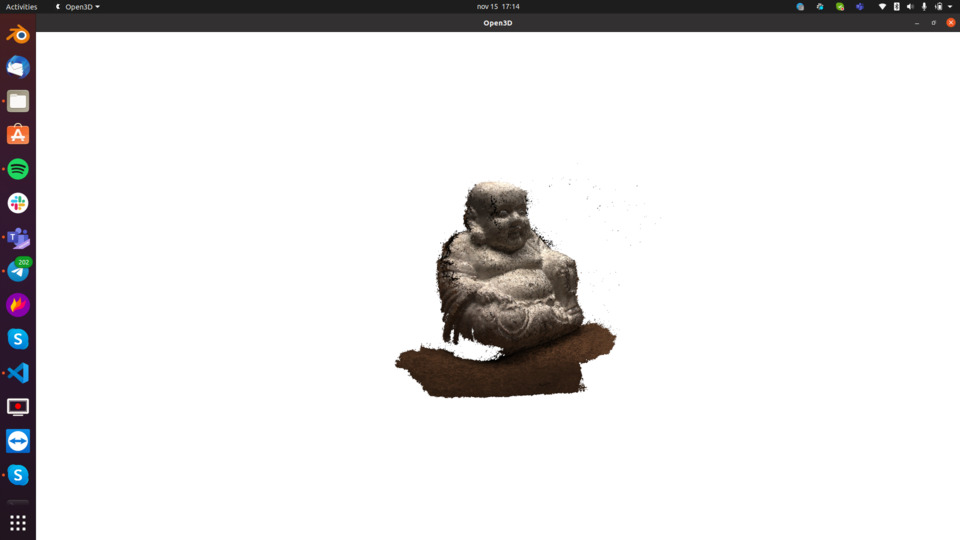} \\
    \end{tabular}
    \caption{\textbf{Qualitative results on DTU dataset -- \textit{scan9} (top) and \textit{scan114} (bottom).}  Depth maps and point clouds yielded by D$^2$HC-RMVSNet (top), CAS-MVSNet (bottom) and guided counterparts trained on BlendedMVG.}
    \label{fig:dtu}
\end{figure*}

\textbf{Coarse-to-fine strategy.} We now ablate the coarse-to-fine guidance mechanism introduced in Sec. \ref{sec:coarsetofine}, by training different variants of CAS-MVSNet. Tab. \ref{tab:coarsetofine} reports results on the BlendedMVG testing split. From top to bottom, we report error rates achieved by the original CAS-MVSNet without guidance, three models guided during one out of the total three stages implemented by the network (i.e. modulating only one out of the three volumes built during inference), and finally the model guided by modulating any single volume. All guided models implement the filtered mvgMVS formulation.
In general, guiding the volume computed only during the first stage already improves the results of the original network. Guiding the second or third stage alone fails at even improving the results by CAS-MVSNet when not guided. Nonetheless, providing a consistent modulation across the three stages allows for the best results.

\subsection{Multi-View Guided MVS networks}

We now evaluate the impact of the mvgMVS framework on the five state-of-the-art networks selected for our experiments. Specifically, we train both the original networks and their counterpart guided employing filtered mvgMVS.

\textbf{Evaluation on BlendedMVG.} We start by evaluating all the networks on the BlendedMVG testing split, collecting the results in Tab. \ref{tab:blended}. By looking at the original networks, we can notice that models implementing coarse-to-fine processing \cite{gu2020cascade,cheng2020deep,wang2021patchmatchnet} result, in general, more accurate compared to MVSNet and D$^2$HC-RMVSNet, achieving about half the error rates with any threshold. This gap is bridged by guiding both with the filtered mvgMVS framework.

Guided counterparts of CAS-MVSNet, UCSNet and PatchMatchNet are further improved too. In particular, CAS-MVSNet-fmvg and UCSNet-fmvg almost halve the error rates at any given threshold, while PatchMatchNet-fmvg benefits from the guidance in minor measure. We ascribe this latter fact to the random initialization performed at the very first stage of PatchMatchNet, left unchanged when implementing its guided counterpart.

\textbf{Generalization to DTU.} We now aim at assessing the impact of the multi-view guided framework on the generalization capacity of the networks to unseen datasets. Purposely, we evaluate the five networks and their guided counterparts on the DTU testing split without fine-tuning on the DTU training split. Tab. \ref{tab:dtu} (a) collects the outcome of this experiment, reporting error metrics on both estimated depth maps (left), as well as on 3D point clouds (right).

By focusing on the former, differently from the experiments on BlendedMVG, we can notice a consistent margin between MVSNet/D$^2$HC-RMVSNet and coarse-to-fine models \cite{gu2020cascade,cheng2020deep,wang2021patchmatchnet} only concerning the number of pixels with error larger than 1 or 2 mm, with mixed results at the increase of the threshold. By looking at guided counterparts, we can appreciate how they always produce much more accurate depth maps, dramatically reducing the error rates. 

Concerning the quality of the reconstructed 3D point cloud, we can observe that coarse-to-fine models achieve both better accuracy and completeness than MVSNet/D$^2$HC-RMVSNet, confirming their effectiveness. Finally, when guided by accumulated depth hints, any network dramatically improves the quality of the fused point clouds, confirming that considerable improvements on single depth maps translates in better 3D reconstructions.

To summarize, this experiment suggests that mvgMVS notably improves the generalization capacity of MVS networks concerning depth maps accuracy and 3D reconstruction quality. Fig. \ref{fig:dtu} shows some qualitative examples.

\textbf{Fine-tuning and evaluation on DTU.} To confirm that the effect of our framework on 3D reconstructions is not limited to generalization scenarios, we fine-tune all the previous networks on the DTU training split and evaluate their performance.
Tab. \ref{tab:dtu} (b) collects results concerning both estimated depth maps (left) and point clouds (right).

Concerning the original networks, we witness a behaviour similar to the one observed in Tab. \ref{tab:dtu} (a), with a margin between coarse-to-fine models and the others, which is consistent only concerning pixels with error larger than 1 mm. Not surprisingly, any network performs better after being fine-tuned, both in terms of depth maps accuracy and point cloud quality.
However, by looking at guided networks, we can notice how their accuracy is further boosted by the fine-tuning phase, with drops of the error rates much higher than those achieved by the original models.

By looking at reconstructed point clouds, for the original networks, we can observe the same trend as in Tab. \ref{tab:dtu}, with coarse-to-fine models generally producing higher quality point clouds. Once again, the more accurate depth maps yielded by mvgMVS correspond to better reconstructions.

To summarize, our experiments highlight that the mvgMVS framework constantly outperforms the original counterpart concerning generalization capability, as well as when data for fine-tuning is available.

\begin{table}[]
    \centering
    \resizebox{0.4\textwidth}{!}{%
    \begin{tabular}{l|r|cccc}
        \specialrule{.2em}{.1em}{.1em}
        \multicolumn{1}{l|}{Network} & Test Hints &  \multicolumn{1}{l}{\textgreater{}1 Px. } & \multicolumn{1}{l}{\textgreater{}2 Px. } & \multicolumn{1}{l}{\textgreater{}3 Px. } & \multicolumn{1}{l}{\textgreater{}4 Px. } \\ 
        \midrule
        MVSNet \cite{yao2018mvsnet} & - & 0.139 & 0.073 & 0.046 & 0.031 \\
        \midrule
        MVSNet-fmvg & 0.03 & \bfseries 0.076 & \bfseries 0.037 & \bfseries 0.023 & \bfseries 0.015 \\
        MVSNet-fmvg & 0.02 & 0.087           & 0.043           & 0.026           & 0.017           \\
        MVSNet-fmvg & 0.01 & 0.109           & 0.054           & 0.033           & 0.022           \\
        
        MVSNet-fmvg & 0.00 & 0.244           & 0.165           & 0.126           & 0.101           \\
        
        \specialrule{.2em}{.1em}{.1em}
    \end{tabular}%
    }
    \centering
    \caption{\textbf{Ablation study -- changing density at testing time.} Validation errors on BlendedMVG \cite{yao2020blendedmvs} testing scans, MVSNet is trained with 3\% hints density and tested with different densities.}
    \label{tab:limitations}
  
    \end{table}

\textbf{Limitations.} Although our experiments highlight the potential of the Multi-View Guided Multi-View Stereo framework, effective on both synthetic and real datasets, our proposal suffers from a limitation that may be important in some environments: networks trained with a specific hints density do not generalize to less dense hints inputs. Specifically, once a guided network has been trained with a fixed density of input depth points, if such density is not guaranteed at the testing time, the performance will drop.
Table \ref{tab:limitations} investigates this behaviour with a further experiment carried out using MVSNet guided with 3\% hints aggregated over the views during training and tested with varying density. We can notice how, by reducing the number of hints, the network performance lowers as well, although still resulting better than the original MVSNet trained without guidance (first row). However, by neglecting the hints at all (last row), the performance dramatically drops below the original MVSNet.
This behaviour highlights that the network itself exploits the hints almost \textit{blindly} when trained with them, losing much accuracy when the hints are not available during deployment, consistently with \cite{Poggi_2019_CVPR}. Future research will explore better training protocols enabling the slightest drop in accuracy in such circumstances. Moreover, the current evaluation is conducted by simulating the availability of depth hints from a sensor. Further experiments with real sensing devices would allow to assess the robustness of the framework to noise in the depth sparse points, as studied in \cite{Poggi_2019_CVPR}.

\section{Conclusion}

In this paper, we have presented a novel framework for accurate MVS depth estimation. Starting from the successes in binocular stereo \cite{Poggi_2019_CVPR}, we extended guided stereo to fully exploit the potential of the multi-view setup by aggregating multiple depth hints acquired from different viewpoints.
Our experiments with five state-of-the-art MVS networks show the effectiveness of our framework, constantly generating much more accurate depth maps and consequently enabling the reconstruction of higher-quality point clouds. This behaviour is consistent either when generalizing from synthetic to real data or after fine-tuning on real images.

\textbf{Acknowledgment.} This work was partially funded by University of Bologna and Ministero dello Sviluppo Economico (MISE) within the Proof of Concept 2020 program.

{\small
\bibliographystyle{ieee_fullname}
\bibliography{egbib}

\begin{thebibliography}{10}\itemsep=-1pt

\bibitem{aanaes2016large}
Henrik Aan{\ae}s, Rasmus~Ramsb{\o}l Jensen, George Vogiatzis, Engin Tola, and
  Anders~Bjorholm Dahl.
\newblock Large-scale data for multiple-view stereopsis.
\newblock {\em International Journal of Computer Vision}, 120(2):153--168,
  2016.

\bibitem{barnes2009patchmatch}
Connelly Barnes, Eli Shechtman, Adam Finkelstein, and Dan~B Goldman.
\newblock Patchmatch: A randomized correspondence algorithm for structural
  image editing.
\newblock {\em ACM Trans. Graph.}, 28(3):24, 2009.

\bibitem{campbell2008using}
Neill~DF Campbell, George Vogiatzis, Carlos Hern{\'a}ndez, and Roberto Cipolla.
\newblock Using multiple hypotheses to improve depth-maps for multi-view
  stereo.
\newblock In {\em European Conference on Computer Vision}, pages 766--779.
  Springer, 2008.

\bibitem{cheng2020deep}
Shuo Cheng, Zexiang Xu, Shilin Zhu, Zhuwen Li, Li~Erran Li, Ravi Ramamoorthi,
  and Hao Su.
\newblock Deep stereo using adaptive thin volume representation with
  uncertainty awareness.
\newblock In {\em Proceedings of the IEEE/CVF Conference on Computer Vision and
  Pattern Recognition}, pages 2524--2534, 2020.

\bibitem{Cheng_2019_CVPR}
Xuelian Cheng, Yiran Zhong, Yuchao Dai, Pan Ji, and Hongdong Li.
\newblock Noise-aware unsupervised deep lidar-stereo fusion.
\newblock In {\em Proceedings of the IEEE/CVF Conference on Computer Vision and
  Pattern Recognition (CVPR)}, June 2019.

\bibitem{courville2017modulating}
Aaron~C Courville.
\newblock Modulating early visual processing by language.
\newblock In {\em NIPS}, 2017.

\bibitem{furukawa2009accurate}
Yasutaka Furukawa and Jean Ponce.
\newblock Accurate, dense, and robust multiview stereopsis.
\newblock {\em IEEE transactions on pattern analysis and machine intelligence},
  32(8):1362--1376, 2009.

\bibitem{galliani2015massively}
Silvano Galliani, Katrin Lasinger, and Konrad Schindler.
\newblock Massively parallel multiview stereopsis by surface normal diffusion.
\newblock In {\em Proceedings of the IEEE International Conference on Computer
  Vision}, pages 873--881, 2015.

\bibitem{Galliani_2015_ICCV}
Silvano Galliani, Katrin Lasinger, and Konrad Schindler.
\newblock Massively parallel multiview stereopsis by surface normal diffusion.
\newblock In {\em Proceedings of the IEEE International Conference on Computer
  Vision (ICCV)}, December 2015.

\bibitem{gu2020cascade}
Xiaodong Gu, Zhiwen Fan, Siyu Zhu, Zuozhuo Dai, Feitong Tan, and Ping Tan.
\newblock Cascade cost volume for high-resolution multi-view stereo and stereo
  matching.
\newblock In {\em Proceedings of the IEEE/CVF Conference on Computer Vision and
  Pattern Recognition}, pages 2495--2504, 2020.

\bibitem{SemiGlobalMatching}
H. {Hirschmuller}.
\newblock Accurate and efficient stereo processing by semi-global matching and
  mutual information.
\newblock In {\em 2005 IEEE Computer Society Conference on Computer Vision and
  Pattern Recognition (CVPR'05)}, volume~2, pages 807--814 vol. 2, 2005.

\bibitem{PENet}
Mu Hu, Shuling Wang, Bin Li, Shiyu Ning, Li Fan, and Xiaojin Gong.
\newblock Penet: Towards precise and efficient image guided depth completion.
\newblock In {\em 2021 IEEE International Conference on Robotics and Automation
  (ICRA)}, pages 13656--13662, 2021.

\bibitem{huang2017arbitrary}
Xun Huang and Serge Belongie.
\newblock Arbitrary style transfer in real-time with adaptive instance
  normalization.
\newblock In {\em Proceedings of the IEEE International Conference on Computer
  Vision}, pages 1501--1510, 2017.

\bibitem{jensen2014large}
Rasmus Jensen, Anders Dahl, George Vogiatzis, Engil Tola, and Henrik Aan{\ae}s.
\newblock Large scale multi-view stereopsis evaluation.
\newblock In {\em 2014 IEEE Conference on Computer Vision and Pattern
  Recognition}, pages 406--413. IEEE, 2014.

\bibitem{GCNet}
Alex Kendall, Hayk Martirosyan, Saumitro Dasgupta, Peter Henry, Ryan Kennedy,
  Abraham Bachrach, and Adam Bry.
\newblock End-to-end learning of geometry and context for deep stereo
  regression.
\newblock {\em CoRR}, abs/1703.04309, 2017.

\bibitem{Kendall_2017_ICCV}
Alex Kendall, Hayk Martirosyan, Saumitro Dasgupta, Peter Henry, Ryan Kennedy,
  Abraham Bachrach, and Adam Bry.
\newblock End-to-end learning of geometry and context for deep stereo
  regression.
\newblock In {\em The IEEE International Conference on Computer Vision (ICCV)},
  Oct 2017.

\bibitem{knapitsch2017tanks}
Arno Knapitsch, Jaesik Park, Qian-Yi Zhou, and Vladlen Koltun.
\newblock Tanks and temples: Benchmarking large-scale scene reconstruction.
\newblock {\em ACM Transactions on Graphics (ToG)}, 36(4):1--13, 2017.

\bibitem{AQuasiDenseSurfaceReconstruction}
M. Lhuillier and L. Quan.
\newblock A quasi-dense approach to surface reconstruction from uncalibrated
  images.
\newblock {\em IEEE Transactions on Pattern Analysis and Machine Intelligence},
  27(3):418--433, 2005.

\bibitem{luo2019p}
Keyang Luo, Tao Guan, Lili Ju, Haipeng Huang, and Yawei Luo.
\newblock P-mvsnet: Learning patch-wise matching confidence aggregation for
  multi-view stereo.
\newblock In {\em Proceedings of the IEEE/CVF International Conference on
  Computer Vision}, pages 10452--10461, 2019.

\bibitem{SparseToDense}
Fangchang Ma and Sertac Karaman.
\newblock Sparse-to-dense: Depth prediction from sparse depth samples and a
  single image.
\newblock In {\em 2018 IEEE International Conference on Robotics and Automation
  (ICRA)}, pages 4796--4803, 2018.

\bibitem{DispNet}
Nikolaus Mayer, Eddy Ilg, Philip Hausser, Philipp Fischer, Daniel Cremers,
  Alexey Dosovitskiy, and Thomas Brox.
\newblock A large dataset to train convolutional networks for disparity,
  optical flow, and scene flow estimation.
\newblock {\em 2016 IEEE Conference on Computer Vision and Pattern Recognition
  (CVPR)}, Jun 2016.

\bibitem{park2018high}
Kihong Park, Seungryong Kim, and Kwanghoon Sohn.
\newblock High-precision depth estimation with the 3d lidar and stereo fusion.
\newblock In {\em 2018 IEEE International Conference on Robotics and Automation
  (ICRA)}, pages 2156--2163. IEEE, 2018.

\bibitem{park2019high}
Kihong Park, Seungryong Kim, and Kwanghoon Sohn.
\newblock High-precision depth estimation using uncalibrated lidar and stereo
  fusion.
\newblock {\em Ieee transactions on intelligent transportation systems},
  21(1):321--335, 2019.

\bibitem{park2019semantic}
Taesung Park, Ming-Yu Liu, Ting-Chun Wang, and Jun-Yan Zhu.
\newblock Semantic image synthesis with spatially-adaptive normalization.
\newblock In {\em Proceedings of the IEEE/CVF Conference on Computer Vision and
  Pattern Recognition}, pages 2337--2346, 2019.

\bibitem{Poggi_2021_ICCV}
Matteo Poggi, Filippo Aleotti, and Stefano Mattoccia.
\newblock Sensor-guided optical flow.
\newblock In {\em Proceedings of the IEEE/CVF International Conference on
  Computer Vision (ICCV)}, pages 7908--7918, October 2021.

\bibitem{Poggi_2019_CVPR}
Matteo Poggi, Davide Pallotti, Fabio Tosi, and Stefano Mattoccia.
\newblock Guided stereo matching.
\newblock In {\em Proceedings of the IEEE/CVF Conference on Computer Vision and
  Pattern Recognition (CVPR)}, June 2019.

\bibitem{SURVEY_STEREO_DEEP}
Matteo Poggi, Fabio Tosi, Konstantinos Batsos, Philippos Mordohai, and Stefano
  Mattoccia.
\newblock On the synergies between machine learning and binocular stereo for
  depth estimation from images: a survey.
\newblock {\em IEEE Transactions on Pattern Analysis and Machine Intelligence},
  pages 1--1, 2021.

\bibitem{Taxonomy_Stereo}
Daniel Scharstein and Richard Szeliski.
\newblock A taxonomy and evaluation of dense two-frame stereo correspondence
  algorithms.
\newblock {\em Int. J. Comput. Vision}, 47(1–3):7–42, Apr. 2002.

\bibitem{schonberger2016pixelwise}
Johannes~L Sch{\"o}nberger, Enliang Zheng, Jan-Michael Frahm, and Marc
  Pollefeys.
\newblock Pixelwise view selection for unstructured multi-view stereo.
\newblock In {\em European Conference on Computer Vision}, pages 501--518.
  Springer, 2016.

\bibitem{schops2017multi}
Thomas Schops, Johannes~L Schonberger, Silvano Galliani, Torsten Sattler,
  Konrad Schindler, Marc Pollefeys, and Andreas Geiger.
\newblock A multi-view stereo benchmark with high-resolution images and
  multi-camera videos.
\newblock In {\em Proceedings of the IEEE Conference on Computer Vision and
  Pattern Recognition}, pages 3260--3269, 2017.

\bibitem{MVSVoxelColoring}
S.M. Seitz and C.R. Dyer.
\newblock Photorealistic scene reconstruction by voxel coloring.
\newblock In {\em Proceedings of IEEE Computer Society Conference on Computer
  Vision and Pattern Recognition}, pages 1067--1073, 1997.

\bibitem{MVSGraphCuts}
Sudipta~N. Sinha, Philippos Mordohai, and Marc Pollefeys.
\newblock Multi-view stereo via graph cuts on the dual of an adaptive
  tetrahedral mesh.
\newblock In {\em 2007 IEEE 11th International Conference on Computer Vision},
  pages 1--8, 2007.

\bibitem{EnergyMinimization}
Richard Szeliski, Ramin Zabih, Daniel Scharstein, Olga Veksler, Vladimir
  Kolmogorov, Aseem Agarwala, Marshall Tappen, and Carsten Rother.
\newblock A comparative study of energy minimization methods for markov random
  fields with smoothness-based priors.
\newblock {\em IEEE Transactions on Pattern Analysis and Machine Intelligence},
  30(6):1068--1080, 2008.

\bibitem{tang2020learning}
Jie Tang, Fei-Peng Tian, Wei Feng, Jian Li, and Ping Tan.
\newblock Learning guided convolutional network for depth completion.
\newblock {\em IEEE Transactions on Image Processing}, 30:1116--1129, 2020.

\bibitem{GuidedNet}
Jie Tang, Fei-Peng Tian, Wei Feng, Jian Li, and Ping Tan.
\newblock Learning guided convolutional network for depth completion.
\newblock {\em IEEE Transactions on Image Processing}, 30:1116--1129, 2021.

\bibitem{CostAggregationMethods}
Federico Tombari, Stefano Mattoccia, Luigi Di~Stefano, and Elisa Addimanda.
\newblock Classification and evaluation of cost aggregation methods for stereo
  correspondence.
\newblock In {\em 2008 IEEE Conference on Computer Vision and Pattern
  Recognition}, pages 1--8, 2008.

\bibitem{Uhrig2017THREEDV}
Jonas Uhrig, Nick Schneider, Lukas Schneider, Uwe Franke, Thomas Brox, and
  Andreas Geiger.
\newblock Sparsity invariant cnns.
\newblock In {\em International Conference on 3D Vision (3DV)}, 2017.

\bibitem{uhrig2017sparsity}
Jonas Uhrig, Nick Schneider, Lukas Schneider, Uwe Franke, Thomas Brox, and
  Andreas Geiger.
\newblock Sparsity invariant cnns.
\newblock In {\em 2017 international conference on 3D Vision (3DV)}, pages
  11--20. IEEE, 2017.

\bibitem{SemantincMVS}
Ali~Osman Ulusoy, Michael~J. Black, and Andreas Geiger.
\newblock Semantic multi-view stereo: Jointly estimating objects and voxels.
\newblock In {\em Proceedings IEEE Conference on Computer Vision and Pattern
  Recognition (CVPR) 2017}, pages 4531--4540, Piscataway, NJ, USA, July 2017.
  IEEE.

\bibitem{wang2021patchmatchnet}
Fangjinhua Wang, Silvano Galliani, Christoph Vogel, Pablo Speciale, and Marc
  Pollefeys.
\newblock Patchmatchnet: Learned multi-view patchmatch stereo.
\newblock In {\em Proceedings of the IEEE/CVF Conference on Computer Vision and
  Pattern Recognition}, pages 14194--14203, 2021.

\bibitem{wang20193d}
Tsun-Hsuan Wang, Hou-Ning Hu, Chieh~Hubert Lin, Yi-Hsuan Tsai, Wei-Chen Chiu,
  and Min Sun.
\newblock 3d lidar and stereo fusion using stereo matching network with
  conditional cost volume normalization.
\newblock In {\em 2019 IEEE/RSJ International Conference on Intelligent Robots
  and Systems (IROS)}, pages 5895--5902. IEEE, 2019.

\bibitem{Wei_2021_ICCV}
Zizhuang Wei, Qingtian Zhu, Chen Min, Yisong Chen, and Guoping Wang.
\newblock Aa-rmvsnet: Adaptive aggregation recurrent multi-view stereo network.
\newblock In {\em Proceedings of the IEEE/CVF International Conference on
  Computer Vision (ICCV)}, pages 6187--6196, October 2021.

\bibitem{wei2021aa}
Zizhuang Wei, Qingtian Zhu, Chen Min, Yisong Chen, and Guoping Wang.
\newblock Aa-rmvsnet: Adaptive aggregation recurrent multi-view stereo network.
\newblock In {\em Proceedings of the IEEE/CVF International Conference on
  Computer Vision}, pages 6187--6196, 2021.

\bibitem{yan2020dense}
Jianfeng Yan, Zizhuang Wei, Hongwei Yi, Mingyu Ding, Runze Zhang, Yisong Chen,
  Guoping Wang, and Yu-Wing Tai.
\newblock Dense hybrid recurrent multi-view stereo net with dynamic consistency
  checking.
\newblock In {\em ECCV}, 2020.

\bibitem{yang2020cost}
Jiayu Yang, Wei Mao, Jose~M Alvarez, and Miaomiao Liu.
\newblock Cost volume pyramid based depth inference for multi-view stereo.
\newblock In {\em Proceedings of the IEEE/CVF Conference on Computer Vision and
  Pattern Recognition}, pages 4877--4886, 2020.

\bibitem{yao2018mvsnet}
Yao Yao, Zixin Luo, Shiwei Li, Tian Fang, and Long Quan.
\newblock Mvsnet: Depth inference for unstructured multi-view stereo.
\newblock In {\em Proceedings of the European Conference on Computer Vision
  (ECCV)}, pages 767--783, 2018.

\bibitem{yao2019recurrent}
Yao Yao, Zixin Luo, Shiwei Li, Tianwei Shen, Tian Fang, and Long Quan.
\newblock Recurrent mvsnet for high-resolution multi-view stereo depth
  inference.
\newblock In {\em Proceedings of the IEEE/CVF Conference on Computer Vision and
  Pattern Recognition}, pages 5525--5534, 2019.

\bibitem{yao2020blendedmvs}
Yao Yao, Zixin Luo, Shiwei Li, Jingyang Zhang, Yufan Ren, Lei Zhou, Tian Fang,
  and Long Quan.
\newblock Blendedmvs: A large-scale dataset for generalized multi-view stereo
  networks.
\newblock In {\em Proceedings of the IEEE/CVF Conference on Computer Vision and
  Pattern Recognition}, pages 1790--1799, 2020.

\bibitem{MC-CNN}
Jure {\v{Z}}bontar and Yann LeCun.
\newblock Stereo matching by training a convolutional neural network to compare
  image patches.
\newblock {\em Journal of Machine Learning Research}, 17(65):1--32, 2016.

\bibitem{GANet}
Feihu Zhang, Victor Prisacariu, Ruigang Yang, and Philip~HS Torr.
\newblock Ga-net: Guided aggregation net for end-to-end stereo matching.
\newblock In {\em CVPR}, pages 185--194, 2019.

\bibitem{ZhaoYiming_IEEE_2021}
Yiming Zhao, Lin Bai, Ziming Zhang, and Xinming Huang.
\newblock A surface geometry model for lidar depth completion.
\newblock {\em IEEE Robotics and Automation Letters}, 6(3):4457--4464, 2021.

\end{thebibliography}
}

\end{document}